\documentclass[twoside]{article}

\usepackage[accepted]{aistats2022}
\usepackage[round]{natbib}
\usepackage{microtype}
\usepackage{graphicx}
\usepackage{subfigure}
\usepackage{booktabs} % for professional tables
\usepackage{algpseudocode}
\usepackage{algorithm}
\usepackage{color}
\usepackage{amsmath}
\usepackage{amsfonts} 

\begin{document}



\twocolumn[

\aistatstitle{Tile Networks: Learning Optimal Geometric Layout for Whole-page Recommendation}

\aistatsauthor{ Shuai Xiao \And Zaifan Jiang \And  Shuang Yang }

\aistatsaddress{Alibaba Group \And  Alibaba Group \And Alibaba Group } ]

\begin{abstract}
 Finding optimal configurations in a geometric space is a key challenge in many technological disciplines. Current approaches either rely heavily on human domain expertise and are difficult to scale. In this paper we show it is possible to solve configuration optimization problems for whole-page recommendation using reinforcement learning. The proposed \textit{Tile Networks} is a neural architecture that optimizes 2D geometric configurations by arranging items on proper positions. Empirical results on real dataset demonstrate its superior performance compared to traditional learning to rank approaches and recent deep models.
\end{abstract}

\section{Introduction}
Finding optimal configurations in a geometric space is a key challenge in many technological disciplines, for example, \emph{drug discovery}  (i.e., optimal configuration of molecules), \emph{material engineering} (i.e., optimal nanostructure), \emph{circuit design} (i.e., optimal component layout), \emph{logistics} and \emph{transportation} (e.g., optimal plan of routes). For these scenarios, sequential selections of members of \textit{inputs} and \textit{positions} are required to generate optimal configurations. Taking chip design for example, we need to iteratively select chip cells from a pool of cells and place them in appropriate positions onto a chip canvas to obtain a chip of high quality in the end as described by ~\cite{mirhoseini2020chip}.
Current practises in these areas have relied heavily on human domain expertise, as a result, these approaches could be \emph{expensive} (i.e., requiring highly specialized skills), \emph{inefficient} (i.e., throughput bottlenecked by the availability of human experts as well as their latency), \emph{ineffective} (i.e., quality depends entirely on the expertise level of human designers and is often hard to control or to further optimize), and \emph{difficult to scale} (i.e., solutions are often case specific, not easily transferable or automatable) as done in works\citep{Liwo5482, Azami2019, Kalinli14, kool2018attention}. The recent work of \cite{Vinyals2016} suggests a promising new direction of using data-driven approach to tackle this challenging problem automatically using machine learning. 
In particular, for problems like Travelling Salesman Problem (TSP), Convex Hull, the output for the algorithm is the permutation of inputs. 

Now we consider a more complex problem: How to place items and arrange those sequential selections of members of inputs over proper positions?  
The \textit{Pointer Networks} can be modified to solve geometric configuration by sequentially selecting a member of inputs and placing it following a predefined layout, such as top to bottom. We argue that this is not the most efficient way as we can relax from the predefined layout and propose the layout automatically using a data-driven manner which leads to better generalization over the human-defined layout. 
To this end, we present \textit{Tile Networks}, a neural architecture for generating optimal configurations in geometric space by proposing elements from inputs and their positions simultaneously. The model follows an encoder-decoder architecture: the encoder is input-invariant, which aggregates input information via message passing; while the decoder iteratively selects elements from inputs and generate layout by keeping memory of the current state of placement and modeling its influence to future placement.

While our method easily applies to high-dimensional scenarios, we focus on a 2D instantiation of the problem under the application of personalized \emph{whole-page optimization} (WPO), a problem of optimization nature that is prevalent in news feeds, social networks, etc.
Specifically, the goal is, for each given user, to recommend a set of items and at the same time design their layout as displayed on a 2-D geometric space (e.g., the screen of a smartphone or tablet, a webpage of an e-commerce site) in such a way that an objective of interest (e.g., user clicks, purchases, dwell time, revisits) is optimized. To simplify things, we consider the setting that is typical in today's internet industry, where the layout follows a predefined grid template (usually defined by user-interface designers) and items are to be arranged into the grid-bounded cells called ``\emph{tiles}". Even for this simplified case, the configurational space is combinatorial, i.e., for $n$ total items (i.e., inventory size) and $k$ tiles (i.e., display-space size), the total number of possible layouts is $O({k \choose n}\times k!)$, which is daunting for any practical setting.

Conventional approaches to recommender system address the problem by estimating the utility of each item, \emph{either} individually \citep{agarwal2019general} \emph{or} as a pair \citep{Rendle09} or $k$-tuple \citep{Yang2011}, by learning a ranking function. Items are then ranked in a descending order according to the estimated utility, and the top-$k$ list is arranged into \textit{tiles} following a predefined order, such as top to bottom. These methods solve the WPO problem approximately, and they often work but only to the extent where (1) items can be considered independent such that the utility of the bundle \emph{as a whole} are decomposable as the sum (or some linear function) of the individual utilities; and (2) the attraction of the whole page depends entirely on the content of these $k$ items, regardless of their layout. In any realistic scenario, these two assumptions are regarded too restrictive and cannot reflect the truthful reality as shown in works~\citep{ding2019whole, wang2016beyond, zhao2018deep, wang2017efficient}. For example, numerous user studies, e.g.,~\cite{JooWon12} have found that there exist complex interaction patterns not only between items but also among the content of the items, their visual appearances and user's attention distribution over the screen. Our work is similar to recent work of~\cite{wang2016beyond} in the sense that the model generates elements from inputs and their layout simultaneously. \cite{wang2016beyond} proposed using deconvolutional neural networks to generate embedding in each \textit{tile}. The outputs of the deconv in \textit{tiles} are mapped to discrete items through computing the distance between output embedding and item embedding.

We design \textit{Tile Networks}, a neural architecture for generating optimal configurations in geometric, and apply it in the context of personalized whole-page layout in recommender systems. 
The \textit{Tile Networks} can be trained using reinforcement learning to optimize the overall quality of the page as a whole (including the selection of items as well as the layout). 
Experiments in a number of diverse settings show that \textit{Tile Networks} are able to find configurations that are better than alternative methods with a notable margin. The contributions of this paper include:
\begin{itemize}
    \item We propose \textit{Tile Networks}, a neural architecture for optimizing 2D geometric configurations such as personalized whole-page layout in recommender systems.
    \item We design a learning framework that solves geometric configurations optimization based on reinforcement learning.
    \item We testify the \textit{Tile Networks} over different settings of datasets and the results demonstrate \textit{Tile Networks} performs better than baselines including \textit{Pointer Networks} and has a good generalization ability.
\end{itemize}

\section{Problem Formulation}
In this section, we formulate the problem of geometric configuration as a combinatorial optimization problem, and propose machine-learning based solutions for it.

\subsection{Geometric Configuration as Combinatorial Optimization}
Given a $d$-dimensional configuration space and $n$ elements, the goal of geometric configuration is to find the optimal placement of those $n$ elements within the configuration space subject to placement constraints. In this paper, we restrict our attention to the special scenario where the configuration is further constrained onto a grid of size $k=\ell^d$ within the configuration space, as is common in many applications. For instance, in the WPO problem, the recommended commodities are placed onto an $\ell^2$ grid each page, where $\ell$ is the number of rows or columns. 

Denoting a set of elements as $I:=[n]:=\{1,\ldots, n\}$, $\Omega:=[\ell]^d$, a configuration space as $C$ can be mathematically defined as a mapping from $I$ to $\Omega$. Under a given reward function $R(C|I)$ that evaluates a configuration's performance under $I$, the geometric configuration can be posed as an optimization problem
\begin{align}\label{eq::yyx_1}
     \max_{\theta} R(C|I).
\end{align}
The nature of geometric configuration optimization is combinatorial, and is NP-hard, for which a common solution is the greedy algorithm. In the WPO problem, this boils down to a repeated procedure where the learner greedily selects the next element to place on the grid in a greedy fashion. Furthermore, the learner can re-arrange the elements already selected at each iteration to improve upon the performance of the vanilla-greedy selection.

\subsection{Combinatorial Optimization with Reinforcement Learning}

Recently, \cite{bello2016neural} proposed to solve combinatorial optimization problems via reinforcement learning, modeling the greedy selection and rearrangement procedures as outcomes of an action policy. More specifically, denote $C(I_t, \Omega_t)$ as the selected element and the arrangement at the $t$-th iteration, a policy refers to a $\theta$-parameterized probability distribution $p_{\theta}(C(I_t,\Omega_t)|C_{t-1})$ over all possible $C(I_t,\Omega_t)$ that assigns a probability to each possible configuration $C$ as follows:
\begin{align}\label{equ:prob}
    P(C| I, \Omega) = \prod_{t=1}^{k} p_{\theta}\big(C(I_t,\Omega_t)|C_{t-1}\big).
\end{align}
With a slight abuse of notations, denote $C(I_{t})$ as the selected element from $I$ at the $t$-th iteration and $C(\Omega_t)$ as the rearrangement, we further have
\begin{align}
    p_{\theta}\big(C(I_t,\Omega_t)|C_{t-1}\big) = p_{\theta}\big(C(I_t)|C_{t-1}\big) p_{\theta}\big(C(\Omega_t)|C_{t-1}\big).
\end{align}
Under this setting, the goal reduces to finding the best policy parameter $\theta$ by solving
\begin{align}\label{equ:obj0}
    \max_{\theta} J(\theta) = \mathbb{E}_{C\sim p} R(C|I).
\end{align}
In the following, we propose to efficiently solve this formulation using \emph{Tile Networks}, based on the design of \emph{Pointer Networks} which we introduce below.

\subsection{Pointer Networks}
\cite{vinyals2015pointer} proposed a novel seq2seq framework called Pointer Networks which uses attention as a pointer to select a member of inputs which are decoded in a sequence as the output. The architecture can be used to tackle problems that the number of target classes in each step of output depends on the variable input sequence, which can't be well handled by existing sequence-to-sequence networks used by~\cite{sutskever2014sequence}. Pointer Networks trained using supervised learning show promising performance on three one-dimensional geometric configuration problems: finding \emph{planar convex hulls}, computing \emph{Delaunay triangulations}, and solving \emph{travelling salesman problems} (TSPs). \cite{nazari2018reinforcement} improved the seq2seq architecture by replacing RNN encoder with feed-forward networks. They are among the first to use reinforcement learning to train the model for geometric configuration optimization. Finally, latest work by \cite{kool2018attention} further enhanced the architecture by using attention based encoders instead of RNNs. Both \cite{nazari2018reinforcement} and \cite{kool2018attention} applied the algorithms to solve \emph{vehicle routing problems} (VRPs). These works have by now focused primarily on the setting where the solution is a sequence (e.g., a route in TSPs/VRPs, ) and it's no need to further consider the placement positions in geometric space. 
This work attempts to extend this line of work for solving geometric configuration optimization in high-dimensional space where the predefined layout is incapable to handle this complexity.

\section{Neural Network Architecture}
In this section, we present the policy parameterization \textit{Tile Networks} for geometric configuration and the motivation behind it.

\textit{Tile Networks} follows an encoder-decoder network architecture where the main purpose of encoder is to project discrete values $I = \{1,\dots, n\}$ and grid-bounded \textit{tiles} $\Omega$ into a hidden representation. Inspired by previous work of ~\cite{kool2018attention} which lets each member of inputs aware the existence of other members during encoding process, message passing is utilized in our encoder and is implemented through self-attention mechanism. 

Different from \textit{Pointer Networks}, our decoder needs to  select a discrete value from $I = \{1,\dots, n\}$ and place it into a \textit{tile} in space $\Omega$. At the same time, the decoder should memorize previously selected values and their layout to explicitly model their influence to future decisions. 

\begin{figure*}[htb]
\centering
\includegraphics[width=.9\textwidth]{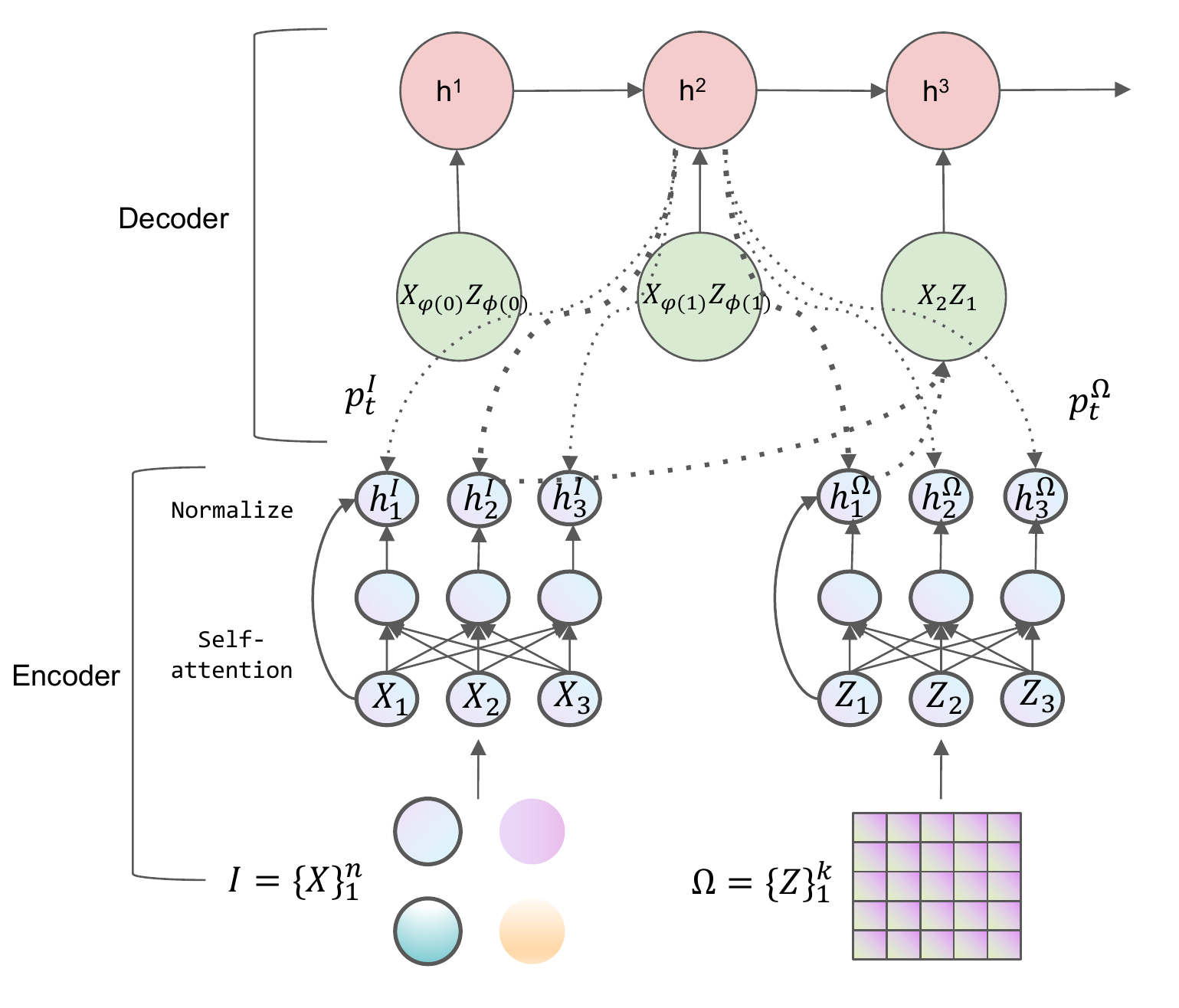}
\caption{The architecture of \textit{Tile Networks}. The dashed arrows pointed to the encoder represent attention over inputs. The dashed arrows pointed to the decoder indicate selecting decision at each decoding step.}
\label{fig:enc-dec}
\end{figure*}

The network architecture is depicted in Figure~\ref{fig:enc-dec}. The parameters of the decoder remain constant regrading to the number of discrete values $I = \{1,\dots, n\}$ and keep relatively compact. 

\textbf{Encoder Network} The discrete values I = \{1,\dots, n\} (e.g., the inventory of items in recommender systems) and \textit{tiles} in geometric space $\Omega$ are encoded separately using self-attention mechanism. For discrete value $i$ of $I$, its feature is represented by a feature vector $X_i $ (which may depend
on users or contexts). For WPO, feature vector $X_i $ contains user features and item features. Thus, the generated pages are conditioned on the personalized user features. 
The $X_i$ acts as query and features $X_1, \ldots, X_n$ of all other values serve as the keys and values simultaneously. Then the hidden state $\mathbf{H}_{i}^{I}$ of discrete value $i$ becomes: 
\begin{align}
\label{attention_affine}
    \mathbf{Q} &= \mathbf{X}\mathbf{W^Q},\quad \mathbf{K} = \mathbf{X}\mathbf{W^K},\quad \mathbf{V} = \mathbf{X}\mathbf{W^V}\\
    \mathbf{H} &= \frac{\text{Softmax}\big(\mathbf{Q} * \mathbf{K}^T)}{\mathbf{\sqrt{D_K}}} \mathbf{V} \quad\\
    \mathbf{H}^{I}  &= \text{LayerNorm}(\mathbf{H} * \mathbf{W^O} + \mathbf{Q}) \label{equ:self-attention-softmax} 
\end{align}
where $\mathbf{Q}, \mathbf{K}, \mathbf{V}$ are query, key, and value matrix after affine transformation by separately multiplying $\mathbf{X}$ with trainable parameters $\mathbf{W^Q}\in\mathcal{R}^{O\times D_{K}}, \mathbf{W^K}\in\mathcal{R}^{O\times D_{K}}, \mathbf{W^V}\in\mathcal{R}^{O\times D_{V}}$. $\mathbf{W^O}$ is the linear projection parameter.

For \textit{tiles} in geometric space $\lambda$, we use $Z_j$ to denote coordinate representation of \textit{tile} $j$. Then the hidden state of \textit{tile} $j$ denoted by $\mathbf{H}_{j}^{\Omega}$ can be obtained using Equation~\ref{attention_affine} and \ref{equ:self-attention-softmax} with a separate set of parameters.

\textbf{Decoder Network} The decoder needs to do two things: one is to select member of inputs and determine their layout; the other is to keep memory of past actions to make future decisions. 
To this end, we use LSTM to keep track of the current state of configuration $C_t$. At step $t$, the hidden state of LSTM decoder is updated as:
\begin{align}\label{equ:dec}
    h^t = \psi(W*h^{t-1} + V*[X_{\varphi(t-1)}, Z_{\phi(t-1)}])
\end{align}
where $h^t$ can be seen as the representation of current configuration $C_t$. $\varphi(t-1)$ is the index of selected discrete value from $I$ at step $t-1$ and $\phi(t-1)$ is the index of selected \textit{tile} at step $t-1$. Therefore, $C(I_t,\Omega_t) =(X_{\varphi(t-1)}, Z_{\phi(t-1)})$ in Equation~\ref{equ:prob}.  $\psi$ is a nonlinear activation function applied element-wise. [] in Equation~\ref{equ:dec} is the vector concatenation operator. Note for the starting step $t=1$, $X_{\varphi(0)}, Z_{\phi(0)}$ are not from inputs but trainable parameters. 
At step $t$, the conditional probability of selecting member $i$ from discrete values $I$ and placing it into \textit{tile} $j$ is:
\begin{align}
    u_i^t &= v^I * \text{tanh}(W_1^I * h^t + W_2^I * H_i^I)\\
    s_j^t &= v^{\Omega} * \text{tanh}(W_1^{\Omega} * h^t + W_2^{\Omega} * H_j^{\Omega})\\
    p_{t}^I &= \text{softmax}(u^t)\label{equ:member}\\
    p_{t}^{\Omega} &= \text{softmax}(s^t)\label{equ:position}
\end{align}
where $\{v^I, W_1^I, W_2^I$ are the trainable parameters of attention networks for selecting members of inputs and $v^{\Omega}, W_1^{\Omega}, W_2^{\Omega} \}$ are parameters of attention networks for designing layout. $p_{t}^I$ is the probability distribution over inputs $I$ and $p_{t}^{\Omega}$ is the probability distribution over \textit{tiles}. Note items and \textit{tiles} that have been selected previously will be masked when computing the probability distributions.

The \textit{Tile Networks} selects a member of inputs and find its placement position sampled from distributions in Equation~\ref{equ:member} and \ref{equ:position} rather than a deterministic way. This stochastic policy allows for exploration of configurations with high rewards.

\textbf{Remark} Although we use whole-page optimization as the testbed, the proposed method \textit{Tile Networks} is not only applicable to the 2D setting and can be directly applied to more general settings, such as genome generation and high-dimensional space, such as 3D, etc. The items and positions can be encoded and decoded in a similar way as in 2D. For the self-attention encoder, the positions are encoded by considering all other positions, similar to text encoders in the natural language process. Therefore, for more general settings, like genome generation or irregular space, each individual position can be encoded only if we know its features, such as coordinates. Our method is not only applicable in regular 2D scenarios. For scenarios with too many possible “tiles”, one possible approach is to aggregate positions into blocks and use the proposed method in a hierarchical manner by iteratively placing items in space.

\section{Learning Algorithms}
\begin{figure*}[htb]
\centering
\includegraphics[width=0.9\textwidth]{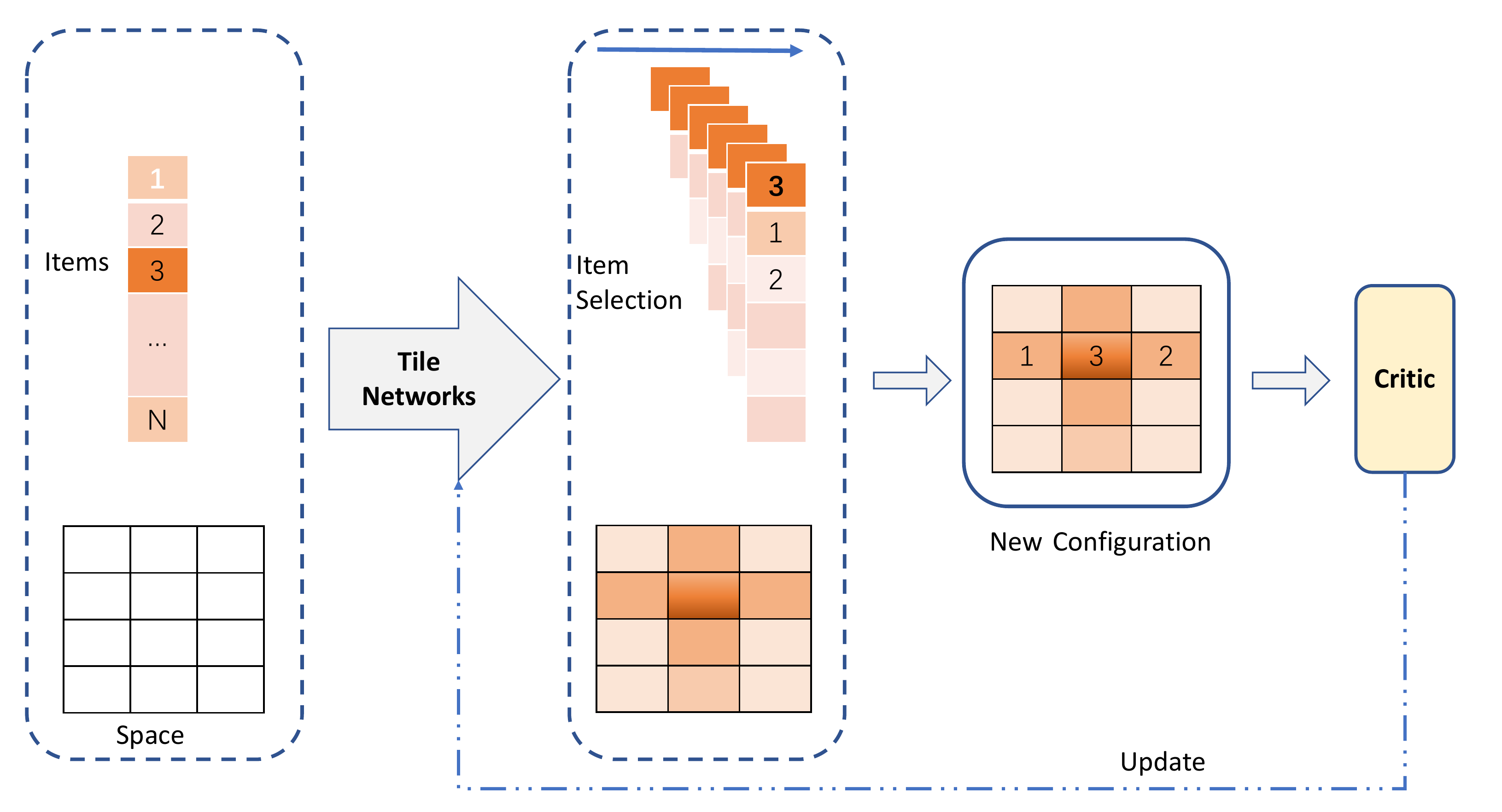}
\caption{The pipeline of learning algorithm. The configurations are sampled from \textit{Tile Networks}, which can be updated through stochastic gradient descent.}
\label{fig:frame}
\end{figure*}

The stochastic policy $p_{\theta}$ parameterized by \textit{Tile Networks} can be optimized through stochastic gradient descent reinforce algorithm proposed by~\cite{williams1992simple}. 
\begin{align}
\nabla_{\theta} J(\theta) = \mathbb{E}_{C\sim p_{\theta}} R(C|I)\nabla_{\theta}\log p_{\theta}(C|I) \label{equ:reinforce}
\end{align}
The reward $R(C|I)$ can be any value with interest
that needs to optimized, such as dwell time, the number of clicks, \emph{NDCG}.
The learning procedure of \textit{Tile Networks} is given in Figure~\ref{fig:frame}. A batch of input samples $\{I\}_1^B$ is draw and their configurations are proposed through the \textit{Tile Networks} and presented to users. The users will decide to click the items based on their preference to the items and viewing order. The reward $R(C|I)$ can be estimated via a critic. Then the policy $p_{\theta}$ is updated using Equation~\ref{equ:reinforce}. The critic is updated by minimizing $\sum_1^B$ $||R(C'|I) - V(C'|I)||^2$ where $V(C'|I)$ is the reward for $C'$ collected from user's feedback. The learning algorithm is summarized in Algorithm~\ref{alg:tilenet}.

 \textbf{Critic Architecture}. The critic has three components: a self-attention encoder, a position-wise fully-connected networks and an aggregate layer. The self-attention encoder is similar to the encoder of \textit{Tile Networks} but the input contains the feature concatenation of $X$ and position embedding $Z$. The position-wise fully-connected networks will apply fully-connected networks to the each hidden states of the encoder. The dimensions for
 queries, keys and values in self-attention is 128. The position-wise fully-connected layer has an output
dimension 16 with ReLU activation function. Then the outputs are concatenated and passed through a fully-connected layer to produce the estimation of expected rewards.

\begin{algorithm}[H]
  \caption{Stochastic policy training of \textit{Tile Networks}}
  \label{alg:tilenet}
  \begin{algorithmic}[1]
    \Require: Training dataset S, batch size B. 
    \Require: $w_0$, initial critic parameters. $\theta_0$,
      initial \textit{Tile Networks}' parameters.
    \While{$\theta$ has not converged}
        \State Sample $\{I\}_1^B$ from training dataset S.
        \State Sample configuration $\{C\}_1^B$ from policy $p_{\theta}(C|I)$.
        \State Estimate reward $R(C|I)$ from critic networks.
        \State Update \textit{Tile Networks}' parameters $\theta$ using Equation~\ref{equ:reinforce}. 
        \State Update critic parameters $w$ 
        by minimizing $\sum_1^B$ $||R(C'|I) - V(C'|I)||^2$.
    \EndWhile
\end{algorithmic}
\end{algorithm}

\section{Experiments}
We conduct experiments on real-world commercial recommendation system datasets. The results show improvement over several core performance metrics.

\noindent\textbf{Datasets and Experimental Setup} The dataset collected from one of the largest e-commercial platforms in the world includes a large-scale records of whole-page recommendations. It contains more than 14 million records with more than 743 thousands users and 7 million items. Both users and items have dense features and categorical features, indicating users' basic profiles and items' features. Users also have personalized features which show users' preferences over different items. Users' click-through data is also recorded where the items in tiles are presented to them. This dataset is publicly available~\footnote{https://github.com/rank2rec/rerank}. We follow the settings in real application scenarios of the whole-page optimization where items are arranged into grid-spaced \textit{tiles} and the tiles' size is 5 rows and 6 columns. 70\% of the entire dataset are randomly sampled for training, 15\% for validation and 15\% for testing.

We adopt the procedure of \cite{joachims2017unbiased} to generate the click data. First of all, we use the recorded click-through data to train a user-item preference model which estimates the click probability of each item when observed by users. Three popular layouts indicating users' scanning behaviors are used, namely 1) top-down row-wise order (Row-Env), 2) left-to-right column-wise order (Col-Env), 3) central area to edges (Z-Env). A real layout (Real-Env) computed from the real dataset is also used. By simulating the user cascade model, $\frac{1}{i^{\eta}}$ represent the observation probability of items observed by users, where $i$ is the item order scanned by users and $\eta=0.05$ is the decay parameter. The click label is generated based on the product of click probability and observation probability.

The model for estimating user-item preference is a feed-forward neural networks with 3 hidden layers and the hidden dimensions and activation functions are (32, 16, 8) and (relu, relu, tanh).  The input is the concatenation of embedding features of users and items. The final layer is a sigmoid function which outputs the click probability. The real scanning order is computed following the method proposed by \cite{position19} by factorizing the click probability into positions' attractiveness and items' attractiveness. The learned positions' attractiveness can be used as the real scanning order. 

To testify the generalization capability of the proposed model, we simulate two item interaction dynamics following the procedure proposed by \cite{bello2018seq2slate}, namely diverse clicks and similar clicks. For diverse clicks, one observed item won't be clicked if it's too similar to previously clicked items. For similar clicks, even an irrelevant item will be clicked when observed if they are very similar to previously clicked items.  The similarity is defined as being the smallest 0.5 percentiles of Euclidean distances between items' features pairs within one page. 

In summary, an environment consists of users' scanning pattern, user-item preference and item interaction dynamic. An environment will simulate user click based on user' viewing priority, items' attractiveness to users and item interaction dynamics. The generated clicks in training dataset are used to train all models. An environment generates clicks for each configuration (layout) output by learned models to evaluate the performance of those models.

\noindent\textbf{Evaluation Metrics} The widely-used metrics in recommendation systems, NDCG and Pre@K are employed to evaluate the model performance. NDCG is normalized discounted cumulative gain which measures the ranking quality. Pre@K calculates the percentage of clicked items in the top K tiles viewed by users. The NDCG/Prec@K are computed following the order of user view behavior, e.g, NDCG is computed in row-wise in Row-Env and in predicted order in Real-Env. The NDCG/Pre@10 computed this way is very similar to the grid metrics
11 used in \cite{xie2019grid} where they first predict user view behavior and then compute grid metrics based on that. The definitions of those metrics are as follows:
\begin{itemize}
    \item \textbf{NDCG} is a measure of ranking quality. NDCG = DCG/IDCG, where DCG is calculated as the following equation:
    \begin{align}
        DCG = \sum_{i=1}^n \frac{2^{rel_i} - 1}{\log_2 (i+1)}
    \end{align}
    where $i$ is the number indicating the item is viewed in the $i$-th order and $rel_i$ is the click label of this item. IDCG is ideal discounted cumulative gain and is equal to DCG after transforming $rel_i$ by sorting $rel_i$ decreasingly. 
    \item \textbf{Pre@K} is defined as the percentage of clicked items in the top-k viewed items. 
    \begin{align}
        Pre@K = \frac{\sum_i^K 1(i)}{K}
    \end{align}
    where $1(i)$ is the indicator function about whether the item in the $t$-th viewed \textit{tile} is clicked.
\end{itemize}

\noindent\textbf{Baselines}
Many sophisticated learning to ranking methods have been developed for personalized recommendation, such as DeepFM ~\citep{guo2017deepfm}, NFM~\citep{he2017neural} and Wide\&Deep model (W\&D)~\citep{cheng2016wide}. Those methods focus on CTR estimation and don’t generate item layout directly. As Wide\&Deep is the most effective recommendation algorithm used in industrial settings, we use Wide\&Deep to represent those methods for CTR estimation. For whole-page recommendation, W\&D model can be used to rank items and then ranked items are placed into \textit{tiles} following a specific layout. The default layout used in practice is top-down row-wise order, which we denote as Row-W\&D model. We also use other two layouts which are widely used in practice which are called Rol-W\&D and Z-W\&D.
\textit{Pointer Networks} proposed by ~\cite{bello2018seq2slate} can be used to determine the output sequence of items. Like W\&D model, we also use three default layouts to place items ranked by \textit{Pointer Networks}, leading to Row-Pointer, Col-Pointer, Z-Pointer model. \cite{wang2016beyond} propose using deconvolutional neural networks (Deconv) to generate optimal configurations, which simultaneously output items and their layout.

For personalized whole-page optimization, the inputs of items' features are the the concatenation of both items' and users' features. For simplicity, the user queries are not explicitly denoted in computation. The items usually contains dense features and categorical features. 

For Wide\&Deep model (W\&D)~\citep{cheng2016wide}, a four-layer fully connected network is used with hidden state size of 128, 64 and activation functions of \emph{ReLU}, \emph{ReLU}. The model is trained using mini-batches of 128 training examples and Adam optimizer with an initial learning rate of 0.001.

For \textit{Pointer Networks}, we follow the implementation as in \cite{bello2018seq2slate} where mini-batches of 128 training examples and LSTM cells with 128 hidden units are used. The optimizer used is Adam with an initial learning rate of 0.001.  

For \textit{Tile Networks}, the self-attention module also contains skip connections~\citep{he2016deep} and layer normalization (LN)~\citep{ba2016layer}. The dimensions of parameters $\mathbf{W^Q}\in\mathcal{R}^{O\times D_{K}}, \mathbf{W^K}\in\mathcal{R}^{O\times D_{K}}, \mathbf{W^V}\in\mathcal{R}^{O\times D_{V}}$  in self-attention network and $\mathbf{W^O}$ are 128. For decoder, the dimensions of LSTM hidden state and  weight matrix $\{v^I, W_1^I, W_2^I, v^{\Omega}, W_1^{\Omega}, W_2^{\Omega} \}$ are all 64. The policy is learned using  ADAM optimizer with an initial learning rate of 0.001. Lastly, our method is implemented in Tensorflow, and the experiments were run with two Nvidia Tesla K80 GPUs.

\begin{figure*}[h]
\centering
\subfigure{
\includegraphics[width=.25\textwidth]{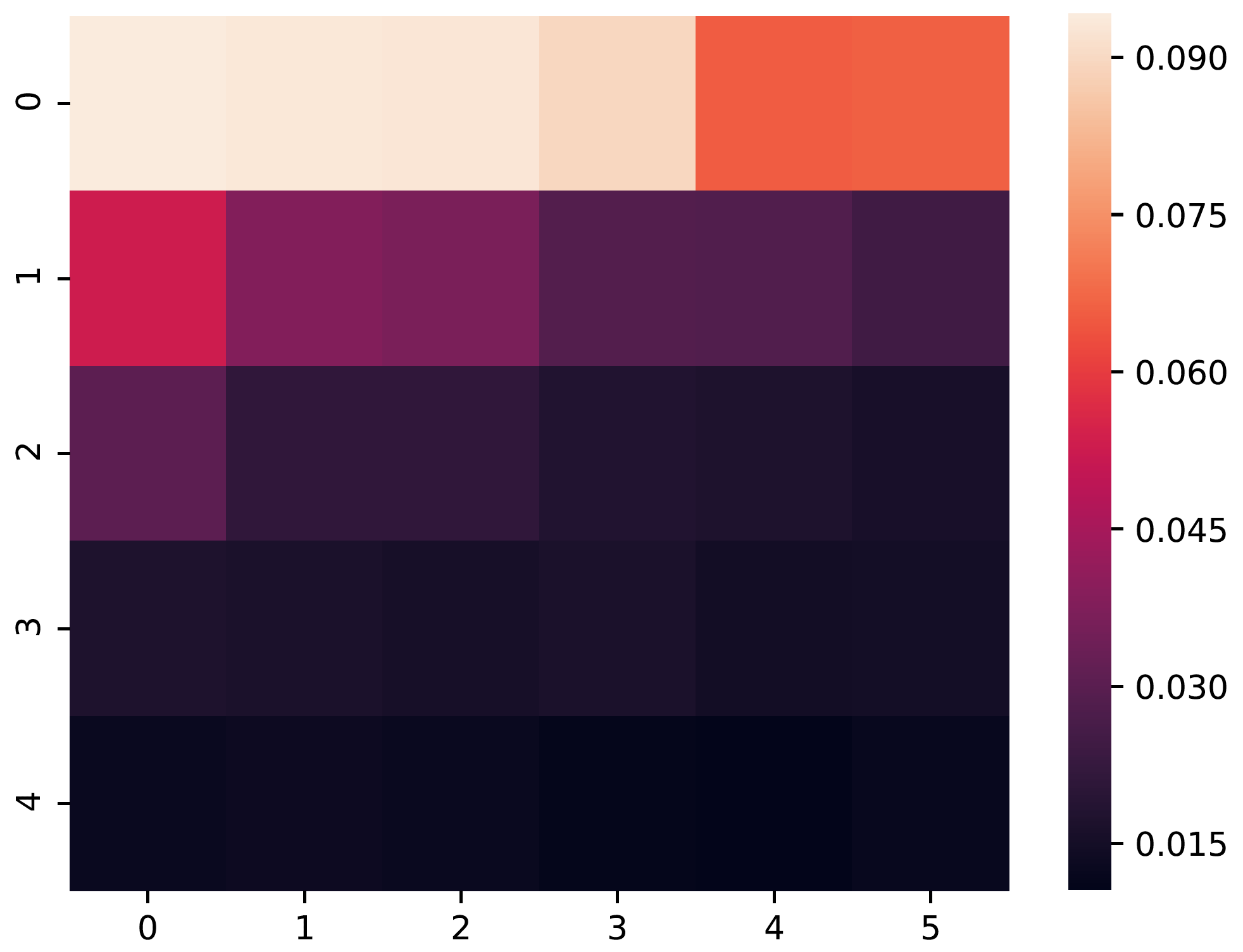}}
\subfigure{
\includegraphics[width=.25\textwidth]{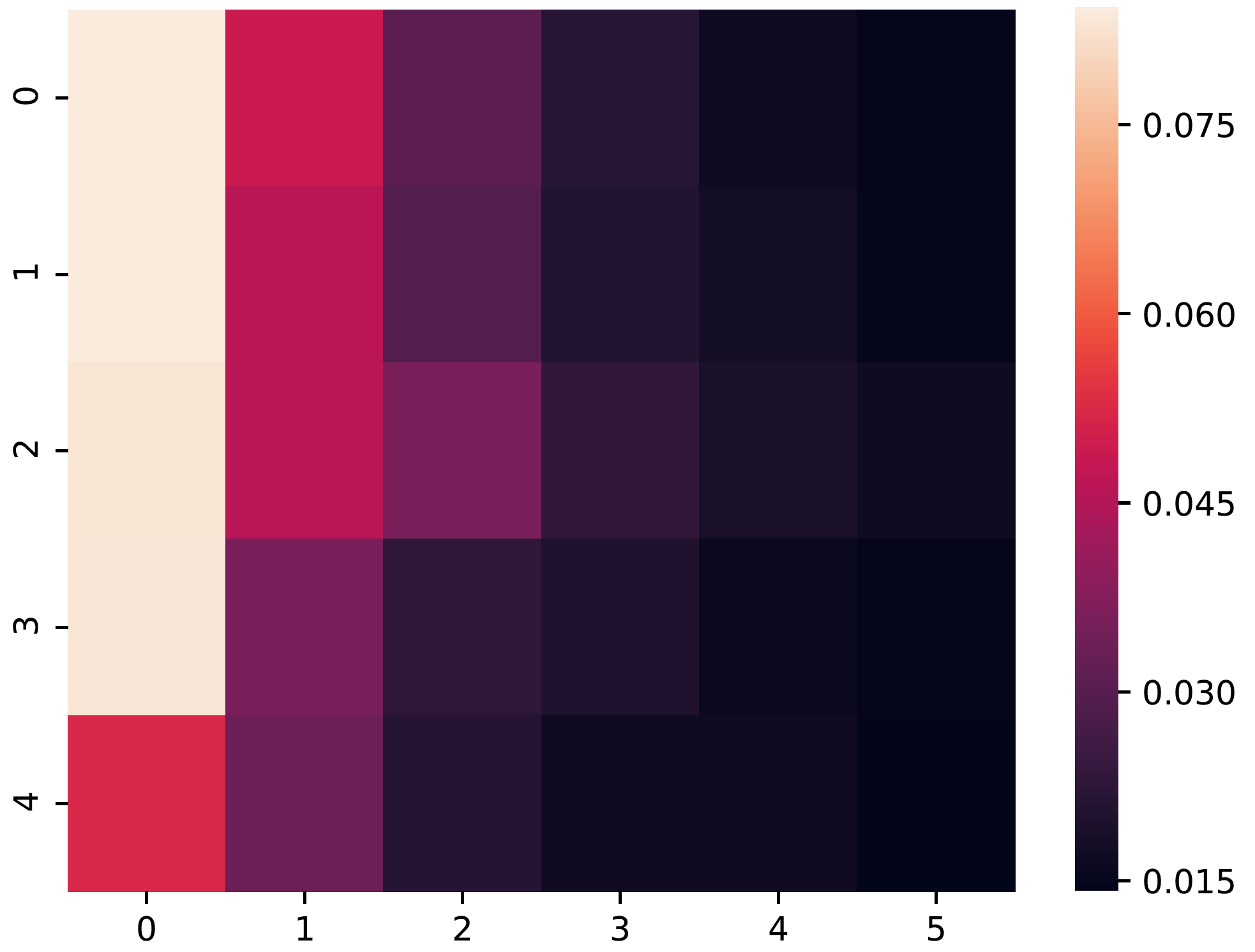}}
\subfigure{
\includegraphics[width=.25\textwidth]{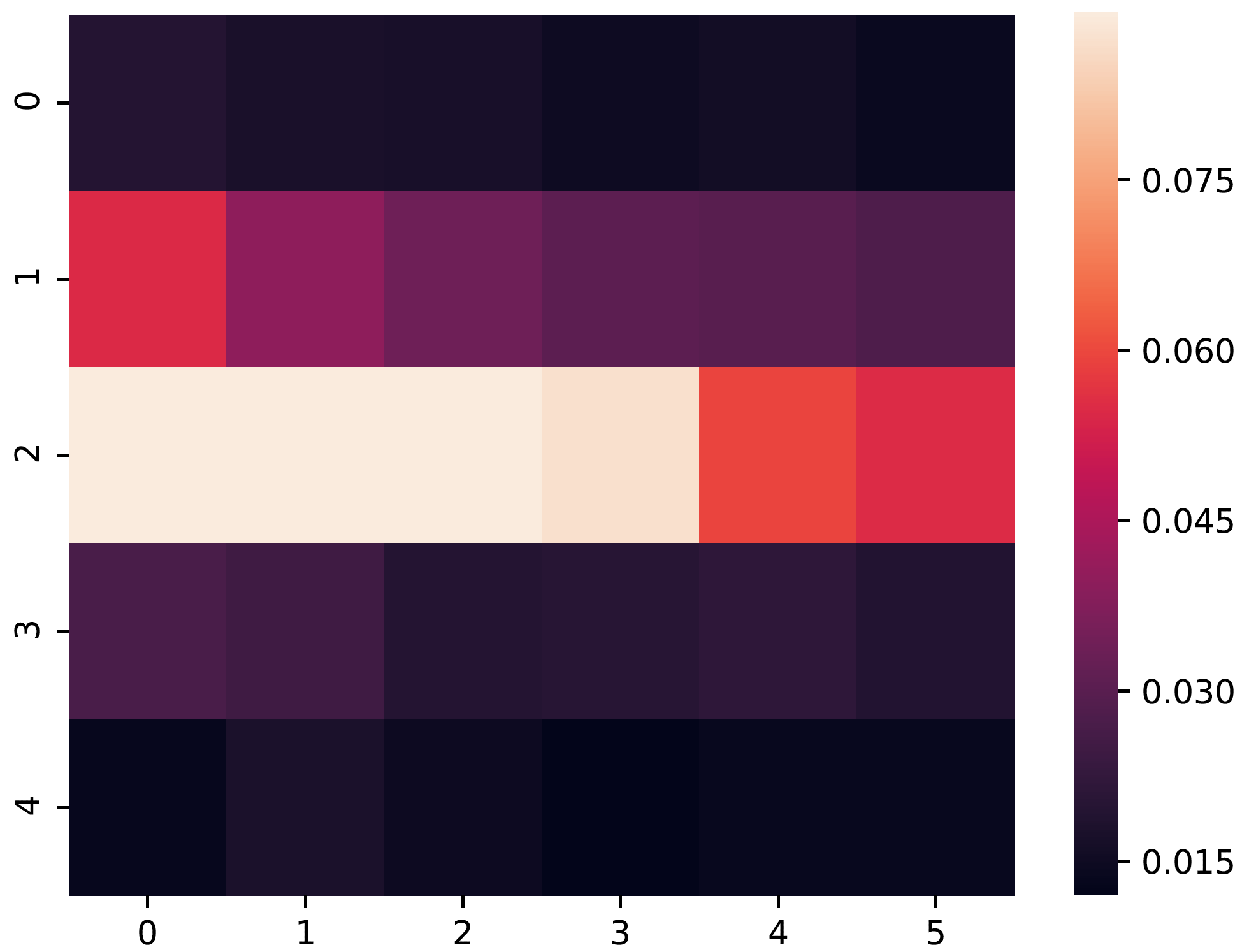}}
\caption{The learned layout for three typical patterns. The left figure shows the learned layout for Row-Env. The middle figure shows the learned layout for Col-Env. The right figure shows the learned layout for Z-Env.}
\label{fig:att}
\end{figure*}

\noindent\textbf{Results}
Table~\ref{table:diverse_result} shows the learning ability of \textit{Tile Networks} when the diversity of items is preferred by users. We can observe that even though \textit{Tile Networks} has to learn the layout, it can still estimate the layout better except for \textit{Pointer Networks} whose default layout coincides with the environment's ground-truth layout. For these cases, \textit{Tile Networks} has a lower but close performance with \textit{Pointer Networks}.  This is quite expected: When learning \textit{Pointer Networks} and we know its optimal layout, the problems reduces to selecting appropriate items. However, for complex layout as in Real-Env, \textit{Tile Networks} outperform \textit{Pointer Networks} with a notable margin. Like \textit{Tile Networks}, Deconv can also select items from inputs and output the layout directly. We can observe that Deconv performs stably across environments but has a lower performance than \textit{Tile Networks}. We speculate this may be due to that Deconv models the interaction of items using local dependency, which can't capture the complex interaction of items. The results also show the  robustness of \textit{Tile Networks} in various settings. 

\begin{table*}[h]
\caption{Models' performance on various environments for diverse clicks.}
\centering
\begin{tabular}{|l |l | l||l |l|| l |l ||l |l |}
 \hline
 & \multicolumn{2}{c||}{Row-Env}& \multicolumn{2}{c||}{Col-Env}&\multicolumn{2}{c||}{Z-Env}&\multicolumn{2}{c|}{Real-Env}\\
 \hline
Algorithm& NDCG &Pre@10 &NDCG &Pre@10 &NDCG &Pre@10& NDCG &Pre@10\\
\hline
\hline
\hline
Row-W\&D     &0.801  &0.172 &0.615  &0.158 &0.697  &0.136 &0.711  &0.152\\
\hline
Col-W\&D      &0.728  &0.159 &0.793  &0.172 &0.708  &0.141 &0.713  &0.154\\
\hline
Z-W\&D      &0.781   &0.160 &0.628  &0.162 &0.732  &0.157 &0.701  &0.141\\
 \hline
Deconv       &0.790  &0.165 &0.658  &0.167 &0.776  &0.171 &0.765  &0.163\\
\hline
Row-Pointer &\bf{0.846}  &\bf{0.191} &0.632  &0.173 &0.613  &0.158 &0.745  &0.161\\
\hline
Col-Pointer &0.652  &0.173 &\bf{0.876}  &\bf{0.186} &0.696 &0.150 &0.736  &0.158\\
\hline
Z-Pointer  &0.684  &0.163 &0.631  &0.165 &\bf{0.827}  &\bf{0.182} &0.721  &0.156\\
 \hline
 \textit{Tile Networks}     &\bf{0.845}  &\bf{0.193} &\bf{0.875}  &\bf{0.186} &\bf{0.826}  &\bf{0.182} &\bf{0.794}  &\bf{0.169}\\
 \hline
\end{tabular}
\label{table:diverse_result}
\end{table*}

\begin{table*}[h!]
\caption{Different models' performance on various environments for similar clicks.} 
\centering
\label{table:similar_result}
\begin{tabular}{|l |l | l||l |l|| l |l || l |l |} 
 \hline
 & \multicolumn{2}{c||}{Row-Env}& \multicolumn{2}{c||}{Col-Env}&\multicolumn{2}{c||}{Z-Env}&\multicolumn{2}{c|}{Real-Env}\\
 \hline
Algorithm& NDCG &Pre@10 &NDCG &Pre@10 &NDCG &Pre@10& NDCG &Pre@10\\
\hline
\hline

\hline
Row-W\&D     &0.803  &0.154 &0.595  &0.129 &0.603  &0.109 &0.672  &0.132\\
\hline
Col-W\&D      &0.751  &0.127 &0.648  &0.150 &0.608  &0.118 &0.683  &0.135\\
\hline
Z-W\&D      &0.728   &0.131 &0.602  &0.137 &0.632  &0.130 &0.670  &0.120\\
 \hline
Deconv       &0.767  &0.143 &0.636  &0.142 &0.629  &0.137 &0.723  &0.143\\
\hline
Row-Pointer &\bf{0.835}  &\bf{0.162} &0.603  &0.130 &0.613  &0.132 &0.712  &0.139\\
\hline
Col-Pointer &0.745  &0.145 &\bf{0.672}  &\bf{0.159} &0.606 &0.127 &0.708  &0.136\\
\hline
Z-Pointer  &0.764  &0.141 &0.618  &0.141 &\bf{0.698}  &\bf{0.152} &0.709  &0.137\\
 \hline
 \textit{Tile Networks}             &\bf{0.831}  &\bf{0.160}    &\bf{0.670}  &\bf{0.158} &\bf{0.695}  &\bf{0.152} &\bf{0.796}  &\bf{0.148}\\
 \hline
\end{tabular}
\end{table*}

To demonstrate the generalization ability and flexibility of \textit{Tile Networks}, we also run models on environments where the concentration of items is preferred by users.
Table~\ref{table:similar_result} shows this empirical results. \textit{Tile Networks} performs better than baselines except for \textit{Pointer Networks} whose default users' viewing priority is the same with that of the environment. This demonstrates that our model can learn from data and is adaptive to various types of interactions in the data.

The learned users' viewing priority (corresponding layouts) are visualized in Figure~\ref{fig:att} after transformation from priority to pixel intensity. The lightness of tiles indicates the \textit{tile} priority of users' viewing of each tiles. The learned priority recover environment ground-truth layout well except small deviations that exist as shown in the Column-wise pattern in Figure~\ref{fig:att}. This uncovered  viewing priority can serve as interpretability map for regions that users pay more attention to. 
Those learned patterns are extremely useful for human–computer interaction study. For instance, promoters can show what they want to sell most in the region where users focus.

The variances of the evaluation metric performance are given in Table~\ref{table:similar_result_variance} to testify the robustness of methods. Tile Networks has comparable variance with Pointer Network, but smaller variance than that of Deconv. 

\begin{table*}[h!]
\caption{Different models' performance variances on various environments for similar clicks.} 
\centering
\resizebox{\linewidth}{!}{%
\label{table:similar_result_variance}
\begin{tabular}{|l |l | l||l |l|| l |l || l |l |} 
 \hline
 & \multicolumn{2}{c||}{Row-Env}& \multicolumn{2}{c||}{Col-Env}&\multicolumn{2}{c||}{Z-Env}&\multicolumn{2}{c|}{Real-Env}\\
 \hline
Algorithm& NDCG &Pre@10 &NDCG &Pre@10 &NDCG &Pre@10& NDCG &Pre@10\\
\hline
\hline

\hline
Row-W\&D     &2.9e-3  &5.3e-3 &8.3e-3 &7.1e-3 &5.9e-3  &4.8e-3 &9.2e-3  &8.5e-3\\
\hline
Col-W\&D      &5.1e-3  &4.6e-3 &8.1e-3  &4.9e-3 &7.3e-3  &6.8e-3 &8.4e-3  &7.9e-3\\
\hline
Z-W\&D      &5.6e-3   &3.2e-3 &5.6e-3  &5.1e-3 &4.1e-3  &3.9e-3 &7.2e-3  &5.7e-3\\
 \hline
Deconv       &4.6e-2  &3.1e-2 &4.5e-2  &4.8e-2 &4.4e-2  &2.7e-2 &5.4e-2  &4.6e-2\\
\hline
Row-Pointer &3.6e-2 &2.9e-2 &3.1e-2  &2.8e-2 &3.8e-2  &2.6e-2 &4.2e-2  &3.7e-2\\
\hline
Col-Pointer &3.9e-2  &3.1e-2 &3.7e-2  &2.7e-2 &2.9e-2 &2.2e-2 &3.7e-2 &3.2e-2\\
\hline
Z-Pointer  &3.4e-2  &2.7e-2 &2.7e-2  &2.1e-2 &3.4e-2  &2.3e-2 &4.2e-2  &3.4e-2\\
 \hline
 \textit{Tile Networks}             &4.2e-2  &3.1e-2   &3.8e-2  &2.8e-2 &3.6e-2  &2.9e-2 &4.4e-2  &3.6e-2\\
 \hline
\end{tabular}
}
\end{table*}

\section{Conclusion}
The paper addresses an under-explored problem: optimal configuration in high-dimensional geometric space. We show a promising direction and propose a neural architecture called \textit{Tile Networks} by extending \textit{Pointer Networks} to solve this problem. To verify the proposed idea, we use 2d personalized whole page optimization as testbed to carry out a series of experiments. Empirical results show that \textit{Tile Networks} which can capture complex item interactions and user’s viewing priority of \textit{tiles} outperforms traditional learning-to-rank methods and recent data-driven approaches. Stable performances on variable environments show its generalization ability. For the future work, designing hierarchical decoding mechanism is a promising direction as the hierarchical structure is more efficient for capturing the interactions both in whole and local scale. Another promising direction is to make the selection of tile dependent on the content which can further improve the model's expressive power. Introducing interpretability into the model is beneficial for improving robustness  and easiness for end users.

\bibliography{example_paper}

%%%%%%%%%%%%%%%%%%%%%%%%%%%%%%%%%%%
%%%%%% SUPPLEMENT (OPTIONAL) %%%%%%
%%%%%%%%%%%%%%%%%%%%%%%%%%%%%%%%%%%

\end{document}